%% file: main.tex
\documentclass{article}

\usepackage{microtype}
\usepackage{graphicx}
\usepackage{subfigure}
\usepackage{booktabs} 
\usepackage{amssymb}
\usepackage{yfonts}
\usepackage{microtype}
\usepackage{hyperref}
\usepackage{setspace}
\usepackage{multirow}

\usepackage[accepted]{icml2021}

\icmltitlerunning{Submission and Formatting Instructions for ICML 2021}

\begin{document}

\twocolumn[
\icmltitle{
Interpreting Criminal Charge Prediction and Its Algorithmic Bias via  Quantum-Inspired Complex Valued Networks}




\begin{icmlauthorlist}
\icmlauthor{Abdul Rafae Khan*}{to}
\icmlauthor{Jia Xu*}{to}
\icmlauthor{Peter Varsanyi}{goo}
\icmlauthor{Rachit Pabreja}{goo}


\end{icmlauthorlist}

\icmlaffiliation{to}{Department of Computer Science, Stevens Institute of Technology}
\icmlaffiliation{goo}{Rutgers University}

\icmlcorrespondingauthor{Jia Xu}{jxu70@stevens.edu}

\icmlkeywords{Deep Learning, Crime Prediction, Interpretability, Bias, Fairness, Quantum Physics, Complex Valued Networks}

\vskip 0.3in
]



\printAffiliationsAndNotice{\icmlEqualContribution} 

\input{sections/abstract.tex}
\input{sections/introduction.tex}

\input{sections/task.tex}

\input{sections/methods.tex}
\input{sections/experiments}

\input{sections/analysis.tex}

\input{sections/relatedwork.tex}

\input{sections/conclusion.tex}

\section*{Acknowledgement}
We appreciate the National Science Foundation (NSF) Award No. 1747728 and the National Science Foundation of China (NSFC) Award No. 61672524 to fund this research. We thank Kovid Inc. to provide the dataset within the NSF project, the initial input of Subhadarshi Panda and the comments of Mengyan Dai, Periklis Papakonstantinou, and Feng Gu. Finally, we sincerely thank the support of the  Google Cloud Research Program.

\bibliography{custom}
\bibliographystyle{icml2021}

\end{document}

%% file: sections/abstract.tex
\begin{abstract}


While predictive policing has become increasingly common in assisting with decisions in the criminal justice system, the use of these results is still controversial. Some software based on deep learning lacks accuracy (e.g., in F-1), and importantly many decision processes are not transparent, causing doubt about decision bias, such as perceived racial and age disparities. This paper addresses bias issues with post-hoc explanations to provide a trustable prediction of whether a person will receive future criminal charges given one's previous criminal records by learning temporal behavior patterns over twenty years. Bi-LSTM relieves the vanishing gradient problem, attentional mechanisms allow learning and interpretation of feature importance, and complex-valued networks inspired quantum physics to facilitate a certain level of transparency in modeling the decision process. Our approach shows a consistent and reliable prediction precision and recall on a real-life dataset. Our analysis of the importance of each input feature shows the critical causal impact on decision-making, suggesting that criminal histories are statistically significant factors, while identifiers, such as race and age, are not. 
Finally, our algorithm indicates that a suspect tends to rather than suddenly increase crime severity level over time gradually. 


\end{abstract}

%% file: sections/introduction.tex
\section{Introduction}

Automatic crime prediction using deep learning algorithms has played an increasingly critical role in criminal justice systems and crime prevention efforts. Previous work, such as~\cite{luo2017learning}, shows that deep learning can assist decision-making by processing a very high volume of data that are difficult for human researchers to analyze efficiently. Despite the success of deep learning, it is still challenging to convince the general public to trust our algorithms to assist a judge's decisions due to a lack of interpretability and sufficient accuracy. 
In particular, criticism that bias exists for identifiers such as race and age points~\cite{skeem2016risk} toward the need for a fair, transparent, and explainable crime prediction method. 
 
Our work presents a criminal charge prediction problem and introduces a framework that leads to an accurate decision with post-hoc interpretability. A criminal charge is a formal accusation made by a governmental authority asserting that an individual has committed a crime. There are three primary classifications of criminal offenses: felonies (we call these level $1$ crimes), including terrorism, murder, kidnapping, treason, elder abuse;  misdemeanors (level $2$), such as arson, extortion, threat, bribery, larceny; and infractions (level $3$), the least severe crimes, including damaging property, burglary, smuggling, obscenity.

We aim to understand individuals' criminal behavior over twenty years and focus our study on whether there are patterns in a criminal's charge record over time. These charge records include a person's prior arrests, type of crime or crimes committed, prior convictions, and personal attributes of race and age. We look specifically at  whether a person will receive a charge in the next two years and if yes, its level given the criminal history in the previous eighteen years and personal information. We also analyze each input feature's importance for prediction results as well as the changes in an individual's crime levels over time.

The first technical challenge in our study is the heavily \textit{imbalanced data}. A fully connected feed-forward neural network can give very low recall and F-measure scores despite high precision. The negative samples are much fewer than the positive samples with the predictor tending to predict everything positive to minimize error. We apply an RNN along with the Bi-LSTM to capture temporal patterns of a suspect's criminal behavior over time, an approach that brings both high precision and high recall.  

The second challenge is the explanation for the causality inference of our predictions. We analyze feature importance using values from attention weights learned from the attentional Bi-LSTMs. 
The weights learned while training and the node connections activated while predicting will indicate the importance of the input features concerning the decision reached during classification.
Our results show that, contrary to the belief that a machine learning algorithm is biased, our algorithms learned that criminal charge histories are essential for prediction and that personal data such as race and age are minimally useful. Our algorithms therefore do not need to explicitly rely on personal information such as race and age that can cause biased decisions. Our result, however, does not indicate whether there is bias in the human raw dataset, a question that is outside the scope of this study. We summarize our three major contributions below:

\begin{enumerate}
    \item We study four research questions on an individual's future criminal charge prediction and introduce Bi-LSTM with attention and Quantum-Inspired Complex Valued Networks (QICVN) to alleviate the imbalanced data problem and to interpret feature importance for decision-making.
    \item Interestingly, our results show that criminal history plays a significant role in predicting a person's future crime and crime level, while personal identifiers, race and age, do not.
        \item Our models indicate that a person is more likely to commit the same or a similar level of crime in the near future than a very different one. 
\end{enumerate}

In the following context, we will first describe our task and dataset, introduce our models, then demonstrate experimental results and interpret our models. Finally, we will discuss related work and conclude.

%% file: sections/task.tex
\section{Task Description and Dataset}\label{sec-task}

\paragraph{Dataset} The real-life crime prediction data we use contains the criminal charge record in Newark, New Jersey, from $1997$ to $2017$, with $17,335$ suspect individuals. 
Each suspect is provided with her/his Personal ID, race, and a list of Bookings. Each booking has a list of sentences, where each sentence has the age of committing the crime and the individual criminal charge history, such as the NCIC crime code, NCIC category code,  and the crime level. The NCIC category code includes the list of ["ASSL","TO", "DAD", "LARC", "FO", "PP", "BURG", "SV", "DP", "OP"]. The input features also include the number of bookings, age average, number of crimes at level $1$, $2$, $3$, and so on. 
We take the first 18 years as the training set and the last two years as the test set. A suspect with at least two bookings is one sample. The label for each sample is the crime level (or whether it is a specific crime level) during the last booking.


\paragraph{Task} 
Our goal is to predict a person's crime or crime level ($1$ for most severe and $3$ least severe) given her/his personal information and previous criminal records. 
We propose a multitude of tasks addressed by 
Question (Q)$1$/$2$/$3$/$4$:

\textit{Is a suspect going to commit a crime at level 1/2/3/any?}

\begin{table}[!t]
    \centering\scalebox{0.8}{
    \begin{small}
    \begin{tabular}{c|c|c||c|c||c|c}
  {\textbf{Age/Race}}& \multicolumn{2}{|c||}{\textbf{Crime level $1$}} & \multicolumn{2}{|c||}{\textbf{Crime level $2$}} & \multicolumn{2}{|c}{\textbf{Crime level $3$}}\\
    \cline{2-7}
     & \textbf{Yes} & \textbf{No} & \textbf{Yes} & \textbf{No} & \textbf{Yes} & \textbf{No} \\     \hline
             All & 6.7& 93.3& 3.5& 96.5&  8.7&91.3\\\hline


    \hline  
                                 
    $\leq$ 20 & 16.9 & 83.1 & 7.2 & 92.8 & 12.8 & 87.2 \\
    21-30 & 7.8 & 92.2 &4.3  & 95.7 & 8.1 & 91.9 \\
    31-50 & 5.9 & 94.1 & 2.9 & 97.1 &9.1 & 90.9 \\
    $>$50 & 2.8 & 97.2 & 1.4 & 98.6 & 7.2  & 92.8 \\
    \hline
    W & 6.8 & 93.2 & 3.6 & 96.4 & 9.2 & 90.8 \\
    A & 9.8 & 90.2 & 2.1 & 97.9 & 8.1 & 91.9 \\
    U & 2.6 & 97.4 & 2.1 & 97.9 & 5.1 & 94.9 \\
    B & 8.6 & 91.4 & 3.6 & 96.4 & 6.9  & 93.1 \\
    I & 3.1 & 96.9 & 0.8 & 99.2 & 6.2  & 93.8 
    \end{tabular}
    \end{small}}
    \caption{Highly imbalanced data: most bookings are labeled as non-crime. We measure the percentage  of the crime- and -non-crime labeled sample number [in \%], respectively. The statistics is calculated on the whole dataset, and on three different data classifications of the age and race, respectively. }
    \label{tab:imbalancestat}
\end{table}

Essentially we train four different systems for each of the above four questions, viewed as a binary classification task. 
We observe from Table~\ref{tab:imbalancestat} that this data is highly imbalanced, with most of the instances having the ``No crime" label.
This resembles a real-life scenario, where a suspect stops committing a crime more often than continuing to commit crimes.
It is technically challenging to handle imbalanced data. 
A fully connected feed-forward neural network does not work in our experiments since it may classify all samples as negative. For example, suppose there are only $2$ positive samples among $100$ samples. In that case, the precision is still $98\%$, although nothing is classified.

We will first describe our task and dataset in the following context, introduce our models, then demonstrate experimental results and interpret our models. Finally, we will discuss related work and conclude.

%% file: sections/methods.tex
\section{Methods} \label{sec_method}

\paragraph{Learning Temporal Patterns with Sequential Models}

To solve the imbalanced data problem, we observe that the criminal charges of an individual follow temporal patterns. A suspect likely repeats a similar behavior, such as committing a crime with the same level or after an equal amount of time (i.e.~somewhat periodically). 
This kind of temporary pattern can be learned with the whole dataset's statistics and shared among different individuals.
We introduce a Recurrent Neural Network (RNN) to capture the sequential patterns of individual criminal charge records along one's lifetime. To reduce the vanishing gradient problem, we add LSTMs~\cite{hochreiter1997long} in the activation functions, which greatly reduced the data imbalance problem. 
As a data preparation step, we had a list of booking for each person. Each booking has an age that shows the person's age when he/she gets the sentence. We find that the maximum number of bookings for a person of different ages (sorted ages from young ages to old ones) is 13. We decided to create sequential data with 12 parts, and we use the last booking as a label of the dataset. 

\paragraph{Interpreting Feature Importance with Attentional Mechanisms}

\textit{The second challenge is that our models lack interpretability. } We provide a certain degree of post-interpretability using the attention mechanism~\cite{vaswani2017attention}.
In particular, the attention mechanism allows to focus on, or pay attention to, different parts of the input sequence to generate a prediction. The attention scores obtained while predicting will indicate the importance of the input features towards the decision and can provide post-hoc interpretability.
To obtain the feature importance: 
First, we check the impact of all the input features on our model predictions by setting the window size to be the total number of input features.
After calculating the embedding of each feature, we compute their relative weights.
Then, we apply softmax overall features to obtain probabilities of the predictions.
Each input is associated with a probability value, and the sum of all probabilities should be $1$.


\paragraph{Increasing Model Transparency with Quantum-inspired Complex Valued Networks}

\begin{figure}[t!]
    \centering
    \includegraphics[scale=0.15]{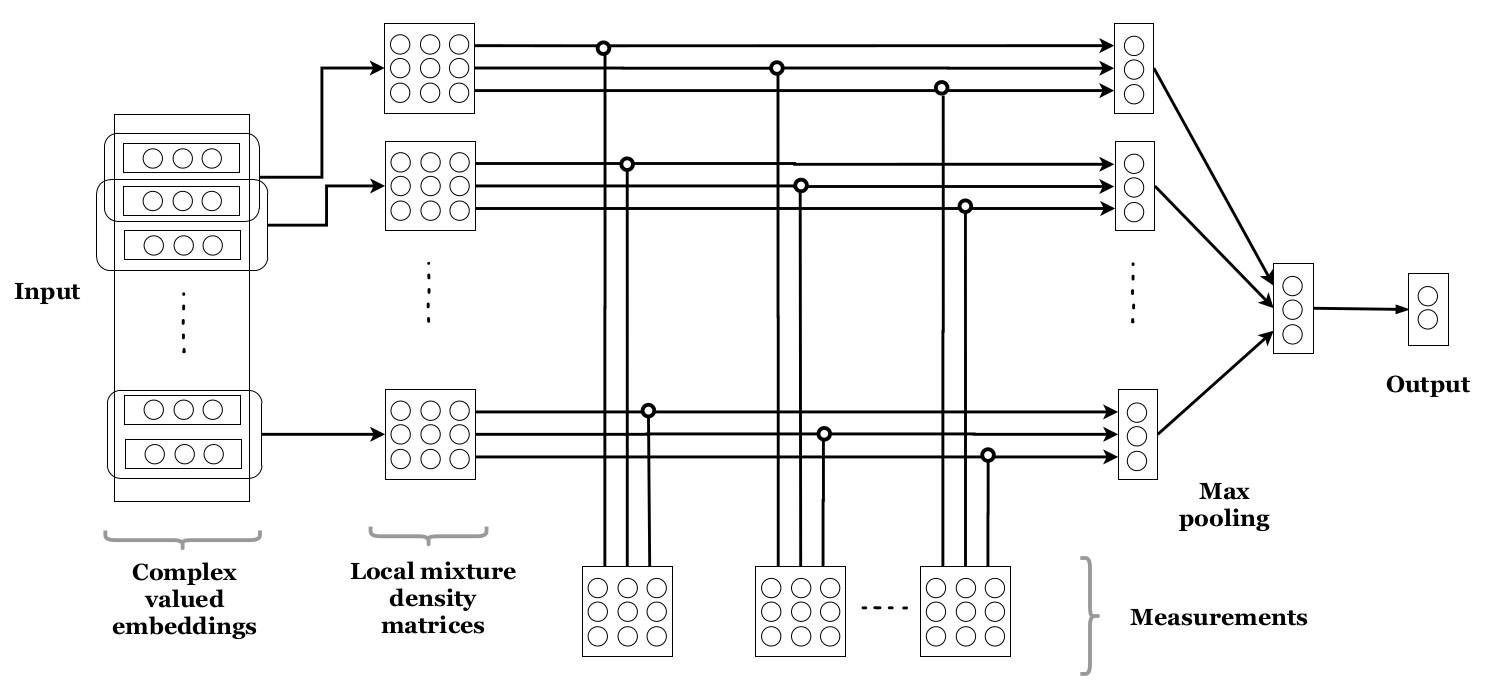}
    \caption{Architecture of Quantum-Inspired Complex Valued Networks (QICVN). Inputs: features; Outputs: whether a person will conduct a criminal charge at a given level. }
    \label{fig:cnm_model}
\end{figure}

Attention mechanisms only interpret the input feature weights, while the deep learning hidden layers and decision process are hard to explain. 
To increase the transparency of our prediction model, we introduce complex-valued networks based on the mathematical formulation of quantum physics. 
As a realization, we adopt the model of~\cite{li2019cnm} from the matching task to a binary classification task. The criminal history and other attributes are the feature inputs to the model, and the output is yes or no to the future crime level.

Specifically,  we first convert each feature value to a complex embedding of size $d$.
 An embedding layer with a ReLU activation \cite{nair2010rectified} takes as input the features one-by-one and converts them into two $d$-dimensional vectors, one for representing amplitude and the other for phase.
 Each number in the phase embedding is normalized to the range $[-\pi, \pi]$.
 Given the complex embedding, we compute the relative weight of the $i$-th feature as the $\ell_2$ norm of its amplitude.
 We then normalize the amplitude by dividing by the $\ell_2$ norm.
 Using the normalized amplitude $A_i$ and the phase $\varphi_i$, we compute the real ($R$) and imaginary ($G$) parts of $i$-th feature as $
     R_i = A_i\cos(\varphi_i)$ and $
     G_i = A_i\sin(\varphi_i)$.
As in Figure~\ref{fig:cnm_model},  we compute the mixture ($M$) for each feature in the n-gram as complex matrix multiplication, and then each mixture is multiplied by a softmax probability of the feature's relative weight in the n-gram. After that, all the mixtures in the n-gram are summed up to obtain the local mixture density matrix. 
The local mixtures are projected using $K$ measurement matrices as in \cite{li2019cnm}. The resulting vectors are max-pooled along each dimension, followed by a linear layer with softmax activation.

%% file: sections/experiments.tex
\section{Experiments}

\paragraph{Setup}
Our model~\cite{winata} is based on long short-term memory (LSTM) network, and Bi-LSTM with attentions \cite{hochreiter1997long} with two hidden layers, where each hidden layer has ten nodes with or without attention layers. The inputs are the features, and the outputs are fed to a linear layer mapping to the target classes. The dropout parameter is set to 0.1. The parameters were updated using the Adam optimizer \cite{kingma2014adam} with a learning rate of 0.001 and random initialization. 

\paragraph{Prediction Results}\label{subsec-result}

\begin{table}[!t]
\begin{center}
\scalebox{0.8}{
    \begin{tabular}{l|l|cccc}
        Model & Crime level &  Acc. & Prec. & Recall & F1 \\\hline
        LSTM & 1 & 94.1 & 92.0 & 94.1 & 92.3 \\
        & 2 & 96.4 & 93.2 & 96.4 & 94.8 \\
        & 3 & 92.3 & 91.2 & 92.3 & 91.5 \\ 
        & Any & 87.5 & 86.8 & 87.5 & 87.0 \\\hline
        Bi-LSTM & 1 & 94.5 & 93.0 & 94.5 & 93.1  \\
        with & 2 & 96.5 & 93.2 & 96.5 & 94.8 \\
        attention & 3 & 91.4 & 87.0 & 91.4 & 87.6 \\
        & Any & 90.6 & 90.5 & 90.6 & 90.5 \\\hline
        QICVN & 1 & 85.2 & 93.9 & 85.2 & 88.5 \\
         & 2 & 85.0 & 95.7 & 85.0 & 89.3 \\
        & 3 & 81.2 & 90.2 & 81.2 & 84.5 \\
        & Any & 86.4 & 90.1 & 86.4 & 87.4 \\
    \end{tabular}}
    \end{center}
    \caption{LSTM, Bi-LSTM with attention, and QICVN results for crime level prediction. }\label{tab:result}
\end{table}

Table~\ref{tab:result}
shows the accuracy scores on the test data using different models on four research questions in Section~\ref{sec-task}. We introduce sequential models, including LSTM, Bi-LSTM with and without attention. 
 We achieve both high precision and high recall overcoming the data imbalance problem by considering the criminal charge records along the previous eighteen years and learning the common patterns across samples.

%% file: sections/analysis.tex
\begin{figure}[t!]
    \centering
    \includegraphics
    [scale=0.4]
    {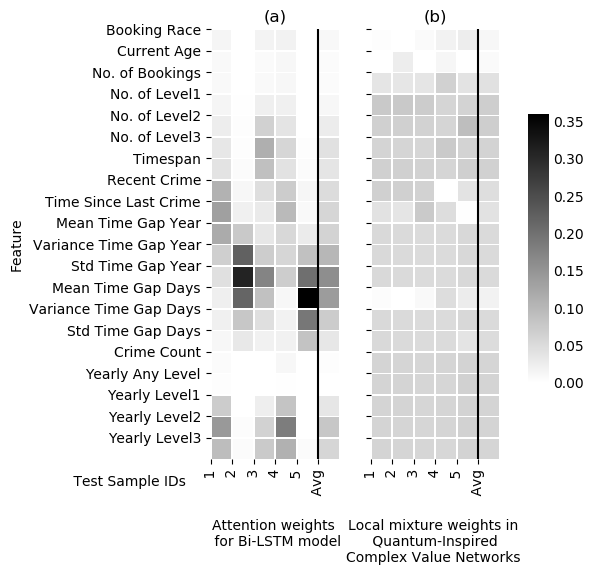}
    \caption{Each column shows the relative weights of different features for one random test sample or the average values across all years. The higher value, the more important is the weight. }
    \label{fig:weights_level_as_class}
\end{figure}

\paragraph{Interpreting Results}\label{subsec-analysis}
Our attention model provides some interpretations for its predictions, suggesting that the prediction relies on historical criminal records rather than personal data, such as race and age.
\textit{Feature weights are learned as a means to measure the  ``causality'' of model decisions in our algorithm.}
We visualize softmax probabilities for randomly selected test samples. See Figure \ref{fig:weights_level_as_class}.
We see that surprisingly, although there are many input features, only a few of them have the highest impact.
Importantly, Figure \ref{fig:weights_level_as_class} shows that personal information such as race and age contribute very little towards the prediction.
Instead, an individual's criminal history features greatly impact the prediction, such as ``time since last crime" and ``variance of the time gap between successive crimes".  
This indicates the personal information features do not explicitly contribute to our algorithmic decisions. Note that our work only focuses on the feature importance explicitly used in our model. Discussions on any bias made by humans in the previous charge decisions~\cite{brantingham2018does,arnold2018racial} are out of the scope of this paper. 



The correlation is computed as the probability (relative frequency) of the lowest level (most severe one) of the crimes committed in the individual history given a certain crime level $L$ in the test set. 
The \textit{algorithmic ``causality''} of level an on predicting level b is computed as follows: For each test sample, we sum up the weights of all features of level a, such as the number of level a, yearly level a,  on the model to predict whether a suspect is going to commit a crime at level b. Then we normalize it with all level-related weights and average the relative weights over all test samples. This average weight shows how much level a influences the decision on predicting level b.

The correlation and causality are different. The former computes the probability of a future crime level given a previous crime level in our data. The latter measures how much the features of a prior crime level contribute to the decision on whether a person will commit a certain level of crime in the future. Thus, the correlation is data-dependent, and the causality is model-dependent. In our study on the transitions of a suspect from one crime level to another, we find that crime level features are essential to predict a new crime level. Table~\ref{fig:transitions} shows whether a lower level crime record indicates a higher level of crime or vice versa. 
Both our algorithmic causality (from our model of Bi-LSTM with attention mechanisms) and correlation results indicate that a suspect is likely to commit a crime with the same or similar level again than a much higher or lower level. 

\begin{table}[!t]
\begin{center}
\scalebox{0.8}{
    \begin{tabular}{l|l|ccc}
        & Crime level (L)& $L=1$ & $L=2$ & $L=3$ \\\hline
    Correlation & $P(\textrm{level } 1|L)$  & 0.53 & 0.37 & 0.25\\
        & $P(\textrm{level } 2|L)$ & 0.15 & 0.33 & 0.19\\
        & $P(\textrm{level } 3|L)$ & 0.32 & 0.30 & 0.56\\\hline
               Causality & level 1 to predict $L$  & 0.60 & 0.12 & 0.11\\
        (Normalized& level 2 to predict $L$ & 0.30 & 0.09 & 0.12\\
        feature weights)& level 3 to predict $L$ & 0.10 & 0.79 & 0.77\\
    \end{tabular}}
    \end{center}
    \label{fig:transitions}
    \caption{Crime level transitions. Given L as the label, the probability/weights of each lowest (most severe crime) level  in history.}
\end{table}

%% file: sections/relatedwork.tex
\section{Related Work}



There has been an increasing trend in criminal justice to leverage machine learning for high-stakes prediction applications that deeply impact human lives.
\cite{6137459} uses Support Vectors Machine (SVM) model
, 2-layered feed forward neural network
and Naive Bayes model 
for crime forecasting.
\cite{stec2018forecasting} and \cite{stalidis2018examining} use feed forward, convolutional and recurrent-convolutional networks, \cite{luo2017learning} uses recurrent based networks with attention mechanism to analyze law articles.
Our model directly predicts the criminal charge based on the criminal history records and interprets our algorithms' decisive factors. ~\cite{dressel2021dangers} courts across the United States are using computer software to predict whether a person will commit a crime, the results of which are incorporated into bail and sentencing decisions. It is imperative that such tools be accurate and fair, but critics have charged that the software can be racially biased, favoring white defendants over black defendants. We evaluate the claim that computer software is more accurate and fairer than people tasked with making similar decisions. We also evaluate and explain the presence of racial bias in these predictive algorithms.

%% file: sections/conclusion.tex
\section{Conclusion}

Our work introduces a trustable criminal charge prediction method with high precision and high recall and post-interpretability. We show that deep learning and quantum-inspired complex-valued networks can be a part of the criminal justice assistant system as long as model transparency and accuracy are taken care of. Perhaps most importantly, we draw attention to the erroneous assumption that social features such as race and age are statistically insignificant to influence our model prediction, even though data may introduce bias. 
